\documentclass[conference]{IEEEtran}
\IEEEoverridecommandlockouts
\usepackage{cite}
\usepackage{amsmath,amssymb,amsfonts}
\usepackage{algorithmic}
\usepackage{graphicx}
\usepackage{textcomp}
\usepackage{xcolor}
\usepackage{booktabs,makecell}
\usepackage[font=small,skip=0pt]{caption}
\usepackage{caption}
\usepackage{subcaption}

\usepackage{adjustbox}

\def\BibTeX{{\rm B\kern-.05em{\sc i\kern-.025em b}\kern-.08em
    T\kern-.1667em\lower.7ex\hbox{E}\kern-.125emX}}
\begin{document}

\title{A deep learning framework to generate realistic population and mobility data\\

\thanks{github/erenarkangil/population-mobility-synthesis}
}

\author{\IEEEauthorblockN{1\textsuperscript{st} Eren Arkangil}
\IEEEauthorblockA{\textit{Deparment of Civil Engineering} \\
\textit{University of Queensland}\\
Brisbane,Australia\\
h.arkangil@uq.edu.au}
\and
\IEEEauthorblockN{2\textsuperscript{nd} Mehmet Yildirimoglu}
\IEEEauthorblockA{\textit{Deparment of Civil Engineering}  \\
\textit{University of Queensland}\\
City, Country \\
email address or ORCID}
\and
\IEEEauthorblockN{3\textsuperscript{rd} Jiwon Kim}
\IEEEauthorblockA{\textit{Deparment of Civil Engineering}  \\
\textit{University of Queensland}\\
City, Country \\
email address or ORCID}

}

\author{
    \IEEEauthorblockN{1\textsuperscript{st} Eren Arkangil\IEEEauthorrefmark{1}, 2\textsuperscript{nd} Mehmet Yildirimoglu\IEEEauthorrefmark{1}, 3\textsuperscript{rd} Jiwon Kim\IEEEauthorrefmark{1}, 4\textsuperscript{th} Carlo Prato\IEEEauthorrefmark{2}}
    \IEEEauthorblockA{\IEEEauthorrefmark{1}The University of Queensland}
    \IEEEauthorblockA{\textit{Deparment of Civil Engineering}
    \\\{h.arkangil, m.yildirimoglu, jiwon.kim\}@uq.edu.au}
    \IEEEauthorblockA{\IEEEauthorrefmark{2}University of Leeds
     \IEEEauthorblockA{\textit{Deparment of Civil Engineering}
    \\\ C.Prato@leeds.ac.uk}}

}

\maketitle

\begin{abstract}
 
Census and Household Travel Survey datasets are regularly collected from households and individuals and provide information on their daily travel behavior with demographic and economic characteristics. These datasets have important applications ranging from travel demand estimation to agent-based modeling. However, they often represent a limited sample of the population due to privacy concerns or are given aggregated. Synthetic data augmentation is a promising avenue in addressing these challenges.  In this paper, we propose a framework to generate a synthetic population that includes both socioeconomic features (e.g., age, sex, industry) and trip chains (i.e., activity locations). Our model is tested and compared with other recently proposed models on multiple assessment metrics.

\end{abstract}


\section{Introduction}

Census and household travel surveys are conducted and published every 5 or 10 years in most countries. They provide rich content about the demographic information of households and geospatial information on the daily trip routines these households have. The Census dataset includes socioeconomic characteristics of the population such as age, sex, and industry, while household surveys include variables related to the personal trip behaviors of the participants in addition to socioeconomic variables. Although Census data provides various demographic features of the households, the information about travel behavior is mostly not given or very limited. On the other hand, household travel surveys are more travel oriented and provides more details in relation to the travel features of the households.

Census and household survey datasets are commonly used to understand and model travel habits and preferences of the households\cite{bindschaedler2017plausible,stopher1996methods,horl2021synthetic}  These datasets are useful for developing traditional travel demand models as well as agent-based simulation models (e.g., MATSim - Multi-Agent Transport Simulation Toolkit). Particularly, the agent-based simulation models require a large amount of micro-level population data to create the agents in the simulation environment. However, the publicly available samples are limited to get comprehensive outcomes 
because they only represent a small proportion of the population. Moreover, published datasets are given aggregated and anonymized due to privacy reasons. For example, the conditional distribution of variables (e.g., age distribution given sex) for the entire population of the census are deemed confidential and usually not published. One of the solutions to overcome these challenges is supporting original samples with synthetic data~\cite{badu2022composite,horl2021synthetic,farooq2013simulation,lederrey2022datgan}.

The process of a synthetic population for transportation systems can be formulated as a two-step process. First, the household population and its socioeconomic features should be created, and afterward, their trip activities need to be combined ~\cite{horl2021synthetic}. Population data synthesis differs from trajectory data synthesis due to the nature of the datasets. Population features are nominal and kept in a tabular format, whereas trajectories are sequential and not necessarily tabular. Although Census data is generally used to generate population data, it is not useful for trajectory prediction and generation tasks. In addition to the socioeconomic features that Census dataset comprises, household travel surveys include activity locations of individuals. Due to their nature, the activity locations are given as a sequence of spatial zones and will be referred to as trip sequences or trajectories.

Our methodology offers a novel way to generate a synthetic population along with corresponding trip sequences. Our framework is capable of generating a relational database including both socioeconomic features and sequential activity locations. Current state-of-the-art models~\cite{badu2022composite,lederrey2022datgan,horl2021synthetic,sun2015bayesian,farooq2013simulation,xu2019modeling} mostly focus on only socioeconomic features, while we offer a broader scope by incorporating sequential trip data into the synthetic population generation problem.

In summary our contributions can be listed as:
\begin{itemize}
\item  Using a Conditional GAN to learn the joint distribution of the socioeconomic variables in the tabular population data
\item  Using recurrent neural networks to create synthetic sequences of locations 
\item  Proposing a novel merging method to integrate population and trip data
\item  Offering new evaluation methods for aggregated and disaggregated results

\end{itemize}

\section{Literature Review}

The proposed algorithms for synthetic data generation problem depends on the type of the data and the task. For example, generative adversarial network models are generally preferred for image generation and recurrent network models for text generation~\cite{shorten2021text}. Our problem includes two types of data: tabular type of population and sequential type of trajectories. The generation of these datasets require different methodologies. There is a wide range of methods that has been proposed to produce synthetic population data; Iterative Proportional Fitting~\cite{horl2021synthetic}, Markov Chain Monte Carlo~\cite{farooq2013simulation}, Bayesian Network~\cite{sun2015bayesian}, and deep learning models~\cite{badu2022composite,lederrey2022datgan}.

Iterative proportional fitting is based on replication of the sample data. It calculates marginal weights for specified columns of the dataset and generates a synthetic population using these weights. IPF based methods are prone to overfitting, because the interactions between columns are not taken into account. Sometimes these weights are available ~\cite{horl2021synthetic}, but it is rarely the case where the weights of all column combinations in this form are available for a given dataset and if the dataset does not contain such information IPF becomes computationally expensive.

Monte Carlo Markov Chain (MCMC) method has been used when direct sampling is difficult and joint distributions $P(x,y)$ are not available. It learns from marginal distributions to generate data, and captures characteristics between variables. The principle of the MCMC approach is to use the Gibbs Sampler and combinations of marginal distributions for up-sampling the empirical data~\cite{farooq2013simulation}. The Gibbs sampler's input is created using marginals and partial conditionals from the dataset. 

Bayesian network (BN) method uses Directed Acyclic Graph (DAG) and structure learning methods to find dependencies between variables. BN does not perform well with complex and large-scale datasets since it is time-consuming and computationally expensive. Additionally, prior knowledge is needed to describe the relation between variables and to build DAGs.

Generative adversarial networks (GAN) are the most up-to-date approach to generate population data. GAN architectures are commonly designed to handle uniform datasets such as generation of images~\cite{karras2020analyzing} rather than tabular datasets. Vanilla and most of the other GAN models suffer from mode collapse problems while generating tabular data~\cite{badu2022composite, xu2019modeling}. Nevertheless, some of the recent GAN models differ from others considering their capability of handling mixed variables (i.e., numerical and categorical) and learning the joint distribution between variables. Of particular interest to our study, conditional GAN ~\cite{xu2019modeling,lederrey2022datgan} models seem to address the limitations that existing algorithms face when generating tabular population data.

While GAN is a promising direction to produce synthetic population data, it has difficulties in directly generating sequences of discrete tokens, such as trajectories or texts. This is partly because taking the gradient of discrete values directly is not possible and using gradient eventually makes little sense~\cite{yu2017seqgan}. Secondly, loss functions used by GANs are not taken into account by partial sequences during generation and score output sequences. Sequence Generative Adversarial Nets with Policy Gradient (SEQGAN)~\cite{yu2017seqgan} adapts the reinforcement learning approach to overcome these problems where generator network is treated as a reinforcement learning agent. This allows the algorithm to generate sequential dataset by using generative networks.

By nature, trajectory generation problem is very similar to text generation problem, where predicting the next element in a sequence is the main task, therefore text generation algorithms can be implemented for trajectory generation purposes. For example, recurrent neural network (RNN) based text generation algorithms are used to create synthetic trajectory data ~\cite{berke2022generating}. Badu-Marfo and Farooq~\cite{badu2022composite} has developed a model named Composite Travel GAN, where the tabular population data is generated by Vanilla GAN and the trip data is generated by SEQGAN. To the best of our knowledge, this is the only study that combines the generation of synthetic population and trajectory problems. Nonetheless, Badu-Marfo and Farooq ~\cite{badu2022composite} does not provide a distinct matching algorithm that enables the merge of the two synthetic datasets, which is one of the contributions that our study makes.

\section{Methodology}

As previously mentioned, most studies focus on either population generation or trajectory generation. However, these two pieces of information are often needed together (e.g., an agent with demographic features and activity locations in an agent-based simulation model). This paper proposes a comprehensive generation framework that handles both tabular population and non-tabular trajectory data. Our framework consists of three main components:  $(i)$ Conditional GAN model to generate synthetic population data, $(ii)$ RNN-based generation algorithm for trip sequence synthesis, $(iii)$ combinatorial optimization algorithm  to merge the output of these two algorithms effectively. Please see Fig. 1 for an overview of the proposed framework. We have two sets of features to be trained. The first set represents population features, and the second represents trip features. First set consists of elements of the population training data $ P_i = \{P_{i_1}, P_{i_2}, P_{i_3}, \dots\ P_{i_m}\} $ where each $P_i$ represents a unique individual and $m$ is the total number of features. The second set is defined as $ L_{i_n} = \{L_{i_1}, L_{i_2}, L_{i_3}, \dots\ L_{i_n} \} $ where i represents individual i and n represents the maximum number of activity locations. Note that the number of activity locations might differ form one individual to another; we consider the individuals that have visited at most n locations. Feature set L is used to train the trajectory generation component of our algorithm.

\begin{table}[!ht]
    \centering
\begin{minipage}[t]{0.48\linewidth}\centering
\caption{Population Features}
\begin{tabular}{ c c }
\toprule
Features &  Samples  \\
\midrule

Age & 25   \\
Sex & Male \\
Industry  & Construction  \\
Origin  &  (-27.468, 152.021) \\
Destination &(-27.270, 153.493)  \\

\bottomrule
\end{tabular}
\end{minipage}\hfill%
\begin{minipage}[t]{0.48\linewidth}\centering
\caption{Sequential Features}
\label{tab:The parameters 2 }
\begin{tabular}{ c c }
\toprule
Features  & Samples  \\
\midrule
Location 1         & Kenmore \\
Location 2           & St. Lucia     \\
Location 3          & South Brisbane    \\
Location 4          & Kenmore  \\
\bottomrule
\end{tabular}
\end{minipage}
\end{table}

Although the feature set P is used to generate population samples, it includes origin destination values in addition to the socioeconomic features. These two features can in fact be matched with the activity locations in the feature set L, see Table 2. These common features will be used later within the bipartite assignment to match the observations in the population and trajectory data, see Figure 1. Nevertheless, origin and destination values in the population feature set P are given in coordinates, while the activity locations in the feature set L are represented as zone names or zone ids. This is because it is computationally easier to train continuous variables in the CTGAN model; the excessive number of discrete zone variables in the network increases the computational complexity and causes memory issues within the CTGAN framework. Therefore, we use centroid coordinates instead of zone ids to train the CTGAN model. We keep the locations as categorical for the feature set L, because sequential models such as RNN and SEQGAN need discrete variables for training process.

\begin{figure}[htb]
    \centering
    \includegraphics[width=8.5cm]{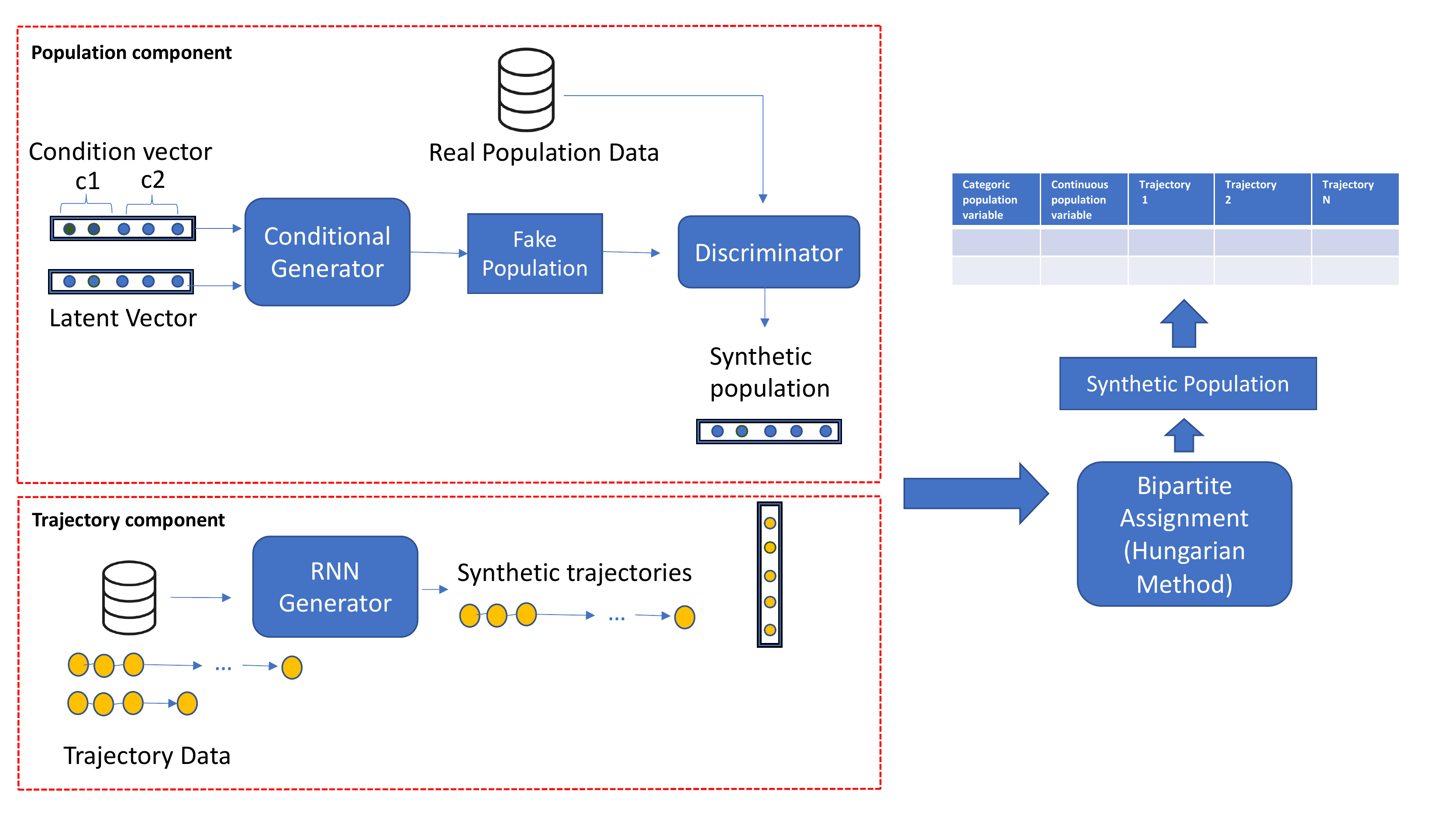}
    \caption{Schematic representation of the framework}
    \label{fig:galaxy}
\end{figure}

\subsection{Tabular data generation}
 
\subsubsection{Generative Adversarial Networks}
Generative Adversarial Networks (GANs) have two neural network components as generator and discriminator, where they play a minimax game against each other. The generator tries to minimize the loss. The discriminator takes input from real and generated data for classification and returns values between 0 and 1 representing the probability of the given input is fake. Therefore, the discriminator wants to maximize the loss.

\begin{equation}
\min_{G}\max_{D}\mathbb{E}_{x\sim p_{\text{data}}(x)}[\log{D(x)}] +  \mathbb{E}_{z\sim p_{\text{z}}(z)}[1 - \log{D(G(z))}]\label{eq}
\end{equation}

The generator produces desired output by using empty latent space and the discriminator behaves like a detective and tries to understand if the output from the generator comes from the real sample or not. 
Latent space is nothing but a feature vector drawn from a Gaussian distribution with a mean of zero and a standard deviation of one. Through training, the generator updates the weights of this vector and learns the real distributions of the sample dataset. Vanilla GAN architecture commonly fails to handle imbalanced and mixed dataset, however recent studies propose significant changes in the GAN architecture to handle tabular data and learn joint distributions~\cite{xu2019modeling,zhao2021ctab,lederrey2022datgan}.

\subsubsection{Conditional GAN}

Conditional Tabular GAN (CTGAN) architecture differs from other GAN models considering its ability to generate tabular data and learn joint distributions. We make use of CTGAN model to synthesize the population because it allows us to generate discrete values conditionally. Besides, it ensures that correleations are kept. CTGAN explores all possible combinations between categorical columns during training. The model consists of fully connected networks to capture all possible relations between columns~\cite{xu2019modeling} and learns probability distributions over all categories. It uses mode-specific normalization and Gaussian mixture to shift the continuous values to an arbitrary range. One hot encoding and Gumble softmax methods are used for discrete values. By using Gumble Softmax trick, the model is able to sample from a categorical variables. The generator creates samples using batch normalization and ReLU (rectified linear unit) activation functions after two fully connected networks. Finally, the Gumble Softmax for discrete values and $tanh$ function is applied to generate synthetic population rows.
 
Besides, it includes PacGAN framework ~\cite{lin2018pacgan} to prevent mode collapse, which is the common failure for vanilla GANs ~\cite{badu2022composite}. CTGAN performs well in synthetic population generation, however, it has difficulties generating realistic trip distributions because the model does not support hierarchical table format and sequential features. Moreover, one hot encoding method can cause memory issues due to high dimensionality of locations. The composite models are proposed to address this issue ~\cite{liu2016coupled,badu2022composite}.

\subsection{Trajectory generation}
 
Synthetic data generation for music, text, and trajectory can be formulated similarly considering the sequential nature of the underlying data. To create input sequences, raw input data is converted into vectors of real numbers. Each discrete value is represented with an unique number in these vectors. This process is called tokenization in natural language processing. The number of tokens is determined by the number of the unique sequence values, which is named vocabulary size in text generation algorithms. RNNs show good performance in predicting and generating conditional elements in a sequence ~\cite{berke2022generating,caccia2018language}. In this study, we use \emph{textgenrnn} library for producing trajectory sequences. The model is inspired by char-rnn~\cite{karpathy2015unreasonable}  and takes trajectory tokens as the input layer. Afterwards, it generates equal size of sequence lists by padding. The sequences are given to the embedding layer as input and create weighted vectors for each token. The embedding layer is followed by two hidden LSTM layers. The attention layer is skip connected to the LSTM and embedding layers. These layers are followed lastly by a dense and output layer.

\subsection{Bipartite assignment}
 
We sample synthetic population and trajectory datasets separately after training CTGAN and RNN models. Because our purpose is to generate individuals with trajectory sequences, we produce the same number of samples for population and trajectory components. We then use the Hungarian algorithm~\cite{munkres1957algorithms} to merge the generated samples.

The Hungarian method takes a non-negative n by n matrix as input, where n represents the number of sampled synthetic individuals. All row and column values are assigned a cost value in the matrix. This cost is defined as the distance between the origin coordinates by CTGAN and RNN components. Since the generated trajectory sequences are discrete, we use the centroid coordinates of the given locations. The cost can be formulated as $d = \sqrt {\left( {x_{ctgan} - x_{rnn} } \right)^2 + \left( {y_{ctgan} - y_{rnn} } \right)^2 }$, where $x_{ctgan}$ and $y_{ctgan}$ are the coordinates of the origin in the synthetic population data, and $x_{rnn}$ and $y_{rnn}$ are the coordinates of the origin zone centroid in the synthetic trajectory data. With the distance values, we build a complete bipartite graph  $G=(P, T)$  with $n$ population vertices $(P$) and $n$ trajectory vertices $(T)$ where the edges are distances between the vertices. Hungarian algorithm finds the optimal matching by minimizing total distance. We use linear assignment without replacement to keep generated trajectories as they are and ensure the trip distributions do not change in the merging process. The algorithm minimizes the total cost (i.e., total distance between the pairs) via the following three steps. Firstly it subtracts the smallest element in each row from all the other elements. Then the smallest entry is equal to 0. Secondly, it subtracts the smallest element in each column. This makes the smallest entry 0 in the column. The steps continue iteratively until the optimal number of zeroes is reached. The optimal number of zeroes is determined by the minimum number of lines required to draw through the row and columns that have the 0 entries. One of the drawbacks of the Hungarian assignment method is the high computation cost with $O(n3)$ time complexity.

\section{Experimental Results}

We test our algorithm using the Queensland Household Travel Survey that includes travel behavior data from randomly selected households. The population and trip activity routines of the households are published as separate tables. Each observation is assigned to a unique person identifier code. We therefore connect population and trip activity tables by using these identifiers. 

The population features contain socioeconomic characteristics such as age, sex, and industry (see Table 1), while the trajectory data includes the sequences of activity locations. The trip locations are categorical and defined as Statistical Area Level 1 (SA1) built from whole Mesh Blocks by Australian Bureau of Statistics. We eliminated observations which contains missing column values and limited the data considering a 40 km radius around Brisbane CBD. We have 5356 unique individuals with their activity chains in our dataset. The maximum number of locations in the trajectory feature set is 4 as 84\% of our observations in training data consist of short trips. In order to build a consistent setup, we train the population and trajectory components with the same dataset. 

\subsection{Parameter Configuration}

We trained CTGAN with 300 epochs with batch size is 10. The layer sizes for the generator and discriminator are (512,512,512). The GAN models are fragile and hard to configure, thus we tested the hyperparameter with different values. We applied the same number of epochs and batch sizes to our RNN model for training. We picked the temperature value as 0.7. The temperature takes values between 0 and 1. It allows the algorithm to generate more divergent results. If the temperature is set as 0, it will make the model deterministic and the algorithm will always generate the same outputs.

\subsection{Results}

We will examine the results in three sections. First, we focus on our results for population generation from the CTGAN algorithm. Second, we discuss the trip sequence distributions resulting from the RNN algorithm.
Last, we focus on the final dataset that results from the merging of the synthetic population and trajectory datasets and analyze the conditional distributions between population and trip specific features. We use a wide range of test metrics considering the most prevalent statistical tests in the transportation literature~\cite{lederrey2022datgan,badu2022composite}.

\subsubsection{Synthetic Population Data}

\begin{figure}[htbp]
\centerline{\includegraphics[width=7cm]{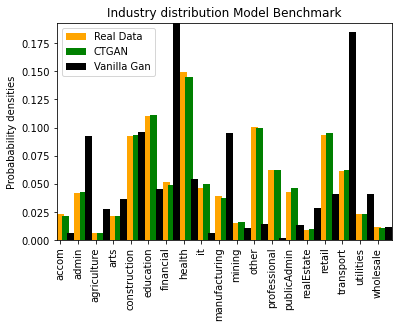}}
\caption{Comparison of marginals for industry attribute}
\label{fig}
\end{figure}

One of the prevalent methods to evaluate synthetic population is comparing population attributes. As discussed in literature review, GAN models are the most successful algorithms to generate synthetic population to our knowledge. We picked Vanilla GAN model for comparison and industry as the feature since amongst other features it has maximum number of classes and therefore hardest to get correct distributions, especially for Vanilla models. As observed in Figure 2, CTGAN outperforms Vanilla GAN in parallel with literature results. Vanilla GAN model cannot learn imbalanced distributions in the tabular data and struggles to reproduce the distribution of categorical industry features.

\subsubsection{Synthetic Trip Sequences}

Comparing trip sequences one at a time only provides little insight into the trip sequence distributions and cannot be used to determine whether the realistic route patterns are formed. Therefore, we consider the trip length to measure the similarity between real and synthetic trip sequences~\cite{badu2022composite,berke2022generating}. We calculate the total distance between all activity locations for each trajectory in the real and synthetic dataset and compare the trip distance distributions that result from this analysis. We also compare the resulting distributions with the synthetic trajectory dataset from SEQGAN that been applied earlier by ~\cite{badu2022composite} for the same purpose. 

Fig 3 clearly shows that RNN closely replicates the trip length distribution in the real dataset, while SEQGAN largely fails to capture the trend in the trip distance distribution. SEQGAN under predicts the distances less than 5km and over predicts the trip lengths greater than 10km, which has been observed in other studies~\cite{badu2022composite} as well. This comparison clearly shows that RNN outperforms SEQGAN and produces realistic trip distance distributions.

\begin{figure}[htbp]
\centerline{\includegraphics[width=8.5cm]{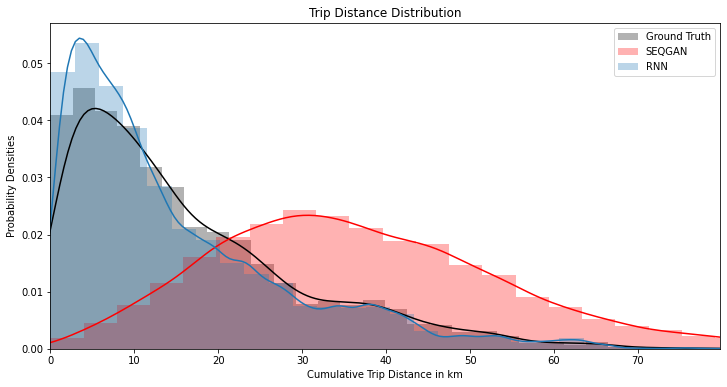}}
\caption{Comparison of trip length distributions}
\label{fig}
\end{figure}

\subsubsection{Synthetic Composite Data}

The marginal distribution of trip distances are solely not enough to show if the correlations are kept between population and trajectories. One needs to consider joint or conditional distributions between population and trip specific features to further examine this aspect. Figure 4 shows conditional trip distances on a given population feature, industry in this case. In particular, Figure 4a presents the distribution of trip distance for those individuals who are employed in the education sector. Note that this conditional distribution can be observed only after the two datasets are merged using the Hungarian algorithm considering the distance between the pairs. Therefore, these results are labeled as 'CTGAN+RNN' in Figure 4a. The comparison between real dataset and the synthetic dataset that results from our algorithm reveals again a very strong similarity. The framework that we propose captures the joint distribution between variables after matching the two datasets.

Figure 4b depicts the distribution of trip distances for all sectors on the coordinate plane to see if the results in 4a are biased only for certain industries. We have 19 industry variables and use 30 bins to obtain trip length intervals. Thus, figure 4b has 570 points in total. Each point on the plane refers to conditional trip distance percentages of the synthetic and real samples. The percentage values on the X and Y axes represent the frequency of the observations in a specific bin interval. Therefore, for each industry, the sum of 19 separate points is equal to 1. Fig. 4b also includes a proportionality line and a regression line to illustrate the overall agreement between synthetic and ground trip distributions.

\begin{figure}[!ht]
 \centering
    \begin{subfigure}[b]{0.4\textwidth}
        \centering
        \includegraphics[width=\textwidth]{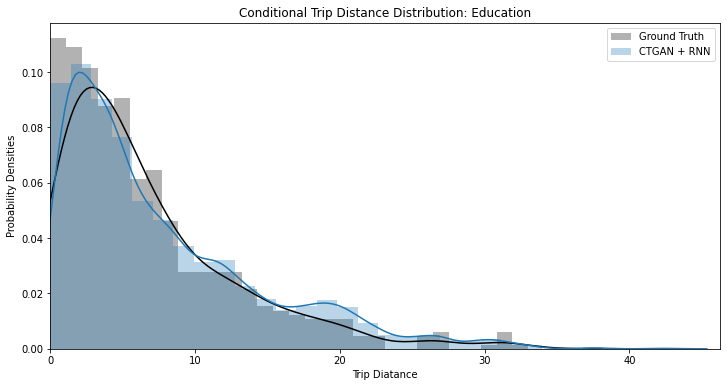}\llap{
  \parbox[b]{3in}{(A)\\\rule{0ex}{1.4in}
  }}
        \label{fig:1a} 
    \end{subfigure}
    ~ 
    \begin{subfigure}[b]{0.4\textwidth}
        \centering
        \includegraphics[width=\textwidth]{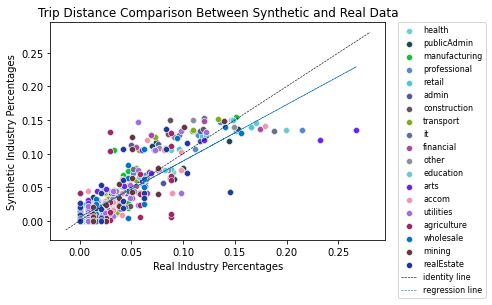}\llap{
  \parbox[b]{3in}{(B)\\\rule{0ex}{1.7in}
  }}
        \label{b}
    \end{subfigure}
    ~ 
    \caption{Comparison of trip length distributions for (a) Education (b) All industries}\label{fig:animals}
\end{figure}

The conditional and unconditional distributions we get from figure 3 and 4 are also tested with statistical metrics; Pearson Correlation, Mean-standardized root mean square error $ (SRMSE)$  and $ R2 $ Score (Coefficient of Determination). We created two frequency lists for the trip length distributions of Conditional GAN, Sequential GAN and RNN models. In Table 3, CTGAN simply indicates the results based on origin and destination values (in coordinates) that are generated as part of the population features. SEQGAN and RNN can be used solely to generate marginal trip distance distributions, while CTGAN+SEQGAN and CTGAN+RNN represents the results after merging of the corresponding population and trajectory datasets. We set the constant term as 0 when we fit a linear regression equation and calculate the $R2$ score. Overall, SEQGAN exhibits a poor performance on imitating real distribution because the generated conditional distributions follow Gaussian distribution rather than tracking the real distribution. Table III gives a summary of the synthetic trip sequence generation performance of three algorithms. 

\begin{table}[htbp]

\def\arraystretch{1}%
\caption{Benchmarking algorithms on statistical tests}
\centering
\begin{adjustbox}{width=\columnwidth,center}
\begin{tabular}{|l|c|c|c|c|}
\hline
\textbf{Trip Distance} & \textbf{Algorithm} & \textbf{Pearson} & \textbf{R2 Score} & \textbf{SRMSE} \\
\hline
 \hspace{0.1cm} & CTGAN  & $0.495$ & $0.231$ & $0.0572$ \\\cline{2-5}
Unconditional & SEQGAN & $ -0.256$  & $ -0.558 $  & $0.116 $ \\\cline{2-5}
 \hspace{0.1cm} & RNN & $0.967$ & $0.927$& $0.0053 $   \\
\hline

\hline

\hline
 \hspace{0.1cm} & CTGAN & $0.302$ & $0.018$ & $0.0479$ \\\cline{2-5}
Conditional & CTGAN + SEQGAN & $ -0.161$  & $  -0.375 $  & $0.0789 $ \\\cline{2-5}
 \hspace{0.1cm} & CTGAN + RNN & $0.867$ & $0.724$& $0.0134$   \\
\hline
\end{tabular} 
\end{adjustbox}

\label{tab:template}
\end{table}

Fig 5 exhibits the spatial distribution of activity locations in both real and synthetic datasets. The most popular activity locations such as airport and city center are coloured with yellow, as most intense areas according to the real samples. As seen in Figure 5, the synthetic location distribution is similar to the ground truth. Even though some of the areas may exhibit a mismatch in terms of trip frequency, the percentage error does not deviate from the real amount significantly, Therefore we can conclude that our method is able to learn geospatial distributions contextually.

\begin{figure}[htbp]

\centering
\includegraphics[width=.45\textwidth]{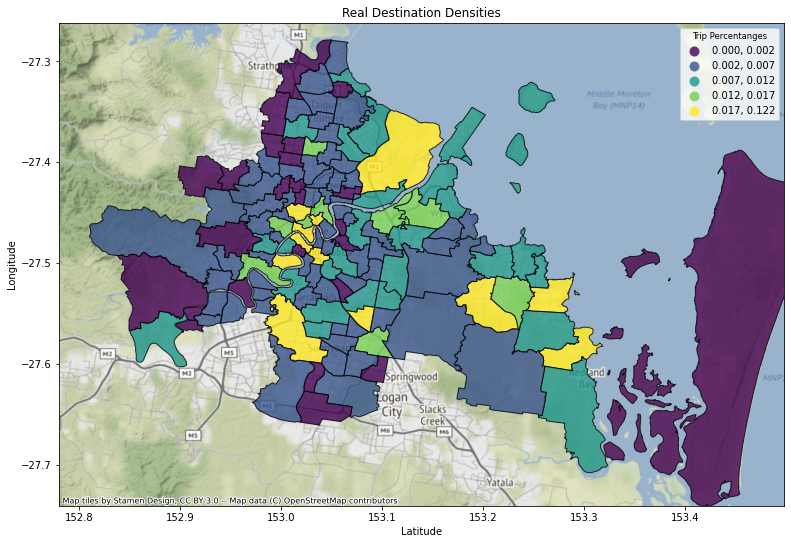}\hfill
\includegraphics[width=.45\textwidth]{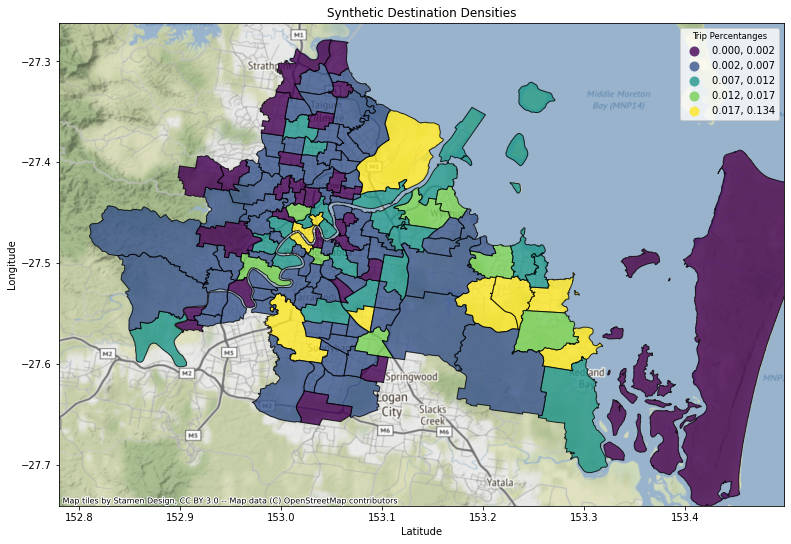}\hfill

\caption{Spatial distribution of real and synthetic datasets}
\label{fig:figure3}

\end{figure}

\section{Conclusion and Future Work}

Synthesis of a mobility database is challenging due to the complex nature of the data. We propose a hybrid framework for generating such complex dataset. We divided the synthesis problem into two sub-parts by generating population and sequential trajectories separately. We use up-to-date models such as CTGAN for tabular data generation and RNN for trajectory generation. We have also applied a combinatorial optimization algorithm to merge these two synthetic datasets. In our future work, we will extend our experiments and design a new framework considering other sources of data to help capture correlations between population and trajectories.

\bibliographystyle{unsrt}
\bibliography{references}

\end{document}